\pdfoutput=1

\documentclass[11pt]{article}

\usepackage{ACL2023}

\usepackage{times}
\usepackage{booktabs}
\usepackage{makecell}
\usepackage{multirow}
\usepackage{latexsym, amsmath, tcolorbox, amssymb}
\usepackage{subfigure}

\usepackage[T1]{fontenc}

\usepackage[utf8]{inputenc}
\usepackage{bbding}
\usepackage{xcolor}
\definecolor{mygray}{rgb}{0.86, 0.86, 0.86}
\usepackage{colortbl}

\usepackage{microtype}

\usepackage{inconsolata}
\usepackage{array}
\newcommand{\cmark}{\textcolor{green}{\CheckmarkBold}}
\newcommand{\xmark}{\textcolor{red}{\XSolidBrush}}
\title{\textit{Filter-then-Generate:} Large Language Models with Structure-Text Adapter for Knowledge Graph Completion}
\newcommand*{\affaddr}[1]{#1} 
\newcommand*{\affmark}[1][*]{\textsuperscript{#1}}
\newcommand*{\email}[1]{\texttt{#1}}

\author{
Ben Liu\affmark[1]\affmark[$\dagger$]\thanks{This work is conducted when Ben Liu was interning at DAMO Academy, Alibaba Group}, Jihai Zhang\affmark[2]\thanks{~~Equal contribution.}, Fangquan Lin\affmark[2],  Cheng Yang\affmark[2], Min Peng\affmark[1]\thanks{~~Corresponding author}\\
\affaddr{\affmark[1]School of Computer Science, Wuhan University, China}\\
\affaddr{\affmark[2]DAMO Academy, Alibaba Group, Hangzhou, 310023, China}\\
\email{\{liuben123, pengm\}@whu.edu.cn}\\
\email{\{jihai.zjh, fangquan.linfq, charis.yangc\}@alibaba-inc.com}
}

\begin{document}
\maketitle
\begin{abstract}
Large Language Models (LLMs) present massive inherent knowledge and superior semantic comprehension capability, which have revolutionized various tasks in natural language processing. Despite their success, a critical gap remains in enabling LLMs to perform knowledge graph completion (KGC). Empirical evidence suggests that LLMs consistently perform worse than conventional KGC approaches, even through sophisticated prompt design or tailored instruction-tuning. Fundamentally, applying LLMs on KGC introduces several critical challenges, including a vast set of entity candidates, hallucination issue of LLMs, and under-exploitation of the graph structure. To address these challenges, we propose a novel instruction-tuning-based method, namely FtG. Specifically, we present a \textit{filter-then-generate} paradigm and formulate the KGC task into a multiple-choice question format. In this way, we can harness the capability of LLMs while mitigating the issue casused by hallucinations. Moreover, we devise a flexible ego-graph serialization prompt and employ a structure-text adapter to couple structure and text information in a contextualized manner. Experimental results demonstrate that FtG achieves substantial performance gain compared to existing state-of-the-art methods. The instruction dataset and code are available at \url{https://github.com/LB0828/FtG}.
\end{abstract}
\section{Introduction}
Knowledge graphs (KGs) encode and store abundant factual knowledge in the format of triples like \textit{(head entity, relation, tail entity)}, which provide faithful knowledge source for downstream knowledge-intensive tasks \citep{kg1, kg2}. However, due to the evolving nature of knowledge, real-world KGs often suffer from incompleteness, urging auto-completion of them. Therefore, knowledge graph completion (KGC), which aims to infer missing triples from the existing KG, has been a fundamental and challenging problem in artificial intelligence research.

\begin{figure}
    \centering
    \includegraphics[width=\linewidth]{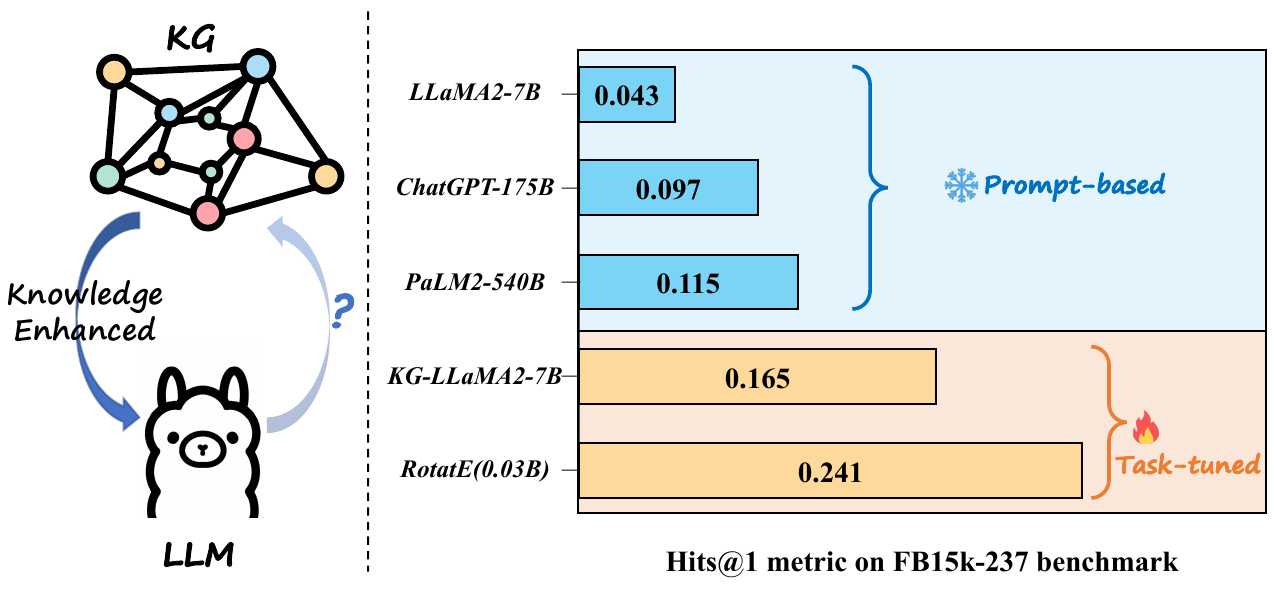}
    \caption{Unsatisfactory performance (Hits@1 metric) of LLMs on the test set of FB15k-237 compared to conventional KGC method RotatE \citep{rotate}.}
    \label{intro}
\end{figure}

Recently, large language models (LLMs) have demonstrated impressive memorization and reasoning abilities through pre-training on massive text corpora \citep{math, bc}. Despite their success, LLMs are limited in insufficient knowledge and prone to generate hallucinations \citep{hall, hall2}. Multiple attempts that exploit KGs to address the hallucinations issues of LLMs have demonstrated promising achievements on various natural language processing tasks \citep{kg2, rag1}. However, the integrating LLMs with KGs for KGC remains underexplored, which present a significant opportunity to revolutionize the approach to KGC by leveraging the contextual understanding and reasoning abilities of LLMs.

\citet{KoPA} and \citet{KG-LLM} apply LLMs to relatively simple KG triple classification (i.e., determining if it is true) has achieved promising results. For more practical and challenging KGC task, \citet{chatgpt1shot} exploit ChatGPT \citep{chatgpt} with in-context learning \citep{in-context} to transform the KGC task into a text-based prediction. However, contrary to expectations, ChatGPT, equipped with hundreds of billions of parameters, lags behind the performance achieved by conventional structure-based methods (e.g., RotatE \citep{rotate}). As shown in Figure~\ref{intro}, on widely-used dataset FB15k-237 \cite{fb15k-237}, ChatGPT only achieves $0.097$ on the Hits@1 metric while RotatE achieves $0.241$. Furthermore, KG-LLaMA \citep{kg-llama} utilizes instruction tuning \citep{instruction-tuning} to adapt LLaMA2-7b \citep{llama2} to KGC task, the performance achieved still is not satisfactory ($0.165$ Hits@1 on FB15k-237). Considering the powerful abilities of LLMs, yet they perform worse than conventional methods on KGC task, a pertinent question arises: \textit{Why LLMs cannot present satisfactory performance on KGC task?}

We speculate LLMs struggle in the KGC task from two aspects: (1) \textbf{Large entity candidate set.} The KGC task can essentially be regarded as a classification task, where the label is all entities in KG. Recent works \citep{rerank} evaluating the performance of LLMs on classification tasks have shown that LLMs struggle in datasets with a large number of labels. The result is consistent with \citet{kg-llama}, where LLMs perform relatively better on the KG with a smaller entity set. The enormity of the entity candidate set in the KG poses a challenge for LLMs. Furthermore, LLMs are known for their propensity to hallucinate content, generating information not grounded by world knowledge \citep{hallucination}. Directly applying LLMs to elicit missing entities from the entire token space often leads to generating invalid entities outside the entity set. (2) \textbf{Inherent Graph Structure of KGs.} Different from text-based NLP tasks, KGs demonstrate unorganized and complicated graph structures. Current efforts \citep{gm} have revealed that performance of LLMs on fundamental graph structural understanding tasks is subpar. Effectively guiding LLMs to comprehend the structural information of KGs remains a considerable challenge.

Building on these findings, we propose a novel instruction-tuning based method, namely FtG, which harnesses and unleashes the capability of LLMs for KGC task. To address the issues of enormous entity candidate set, we present a \textit{filter-then-generate} paradigm, where FtG first employs a conventional KGC method as the filter to eliminate unlikely entities and retain only the top-$k$ candidates. Then we formulate the KGC task into a multiple-choice question format and construct instruction template to prompt LLMs to generate target entity from the top-$k$ candidates. Essentially, this paradigm mirrors human behavior. For example, when answering a question, humans would eliminate obviously wrong answers and find the answer from few remaining candidates. In this way, we narrow the candidate set significantly, and the multiple-choice question format effectively avoids LLMs to output uncontrollable text. Moreover, to incorporate graph structure information into LLMs, we devise a structure-aware ego-graph serialization prompt and propose a light-weight structure-text adapter to map the graph features into the text space. Comparison experiment results show that our proposed FtG greatly improves performance of LLMs on KGC task. In summary, our contributions are:
\begin{itemize}
    \item We propose a novel instruction-tuning based method FtG to enhance the performance of LLMs in the KGC task. In which, our proposed \textit{filter-then-generate} paradigm can effectively harness the powerful capability of LLMs while mitigating the issues caused by hallucinations.
    
    \item To bridge the gap between the graph structure and LLMs, we introduce a structure-aware ego-graph prompt and devise a structure-text adapter in a contextualized way.
    
    \item Our FtG outperforms the state-of-the-art methods over three widely used benchmarks, which demonstrates our model's superiority. Further results reveal that our FtG paradigm can enhance existing KGC methods in a plug-and-play manner. 
\end{itemize}

\begin{figure*}[t]
    \centering
    \includegraphics[width=\linewidth]{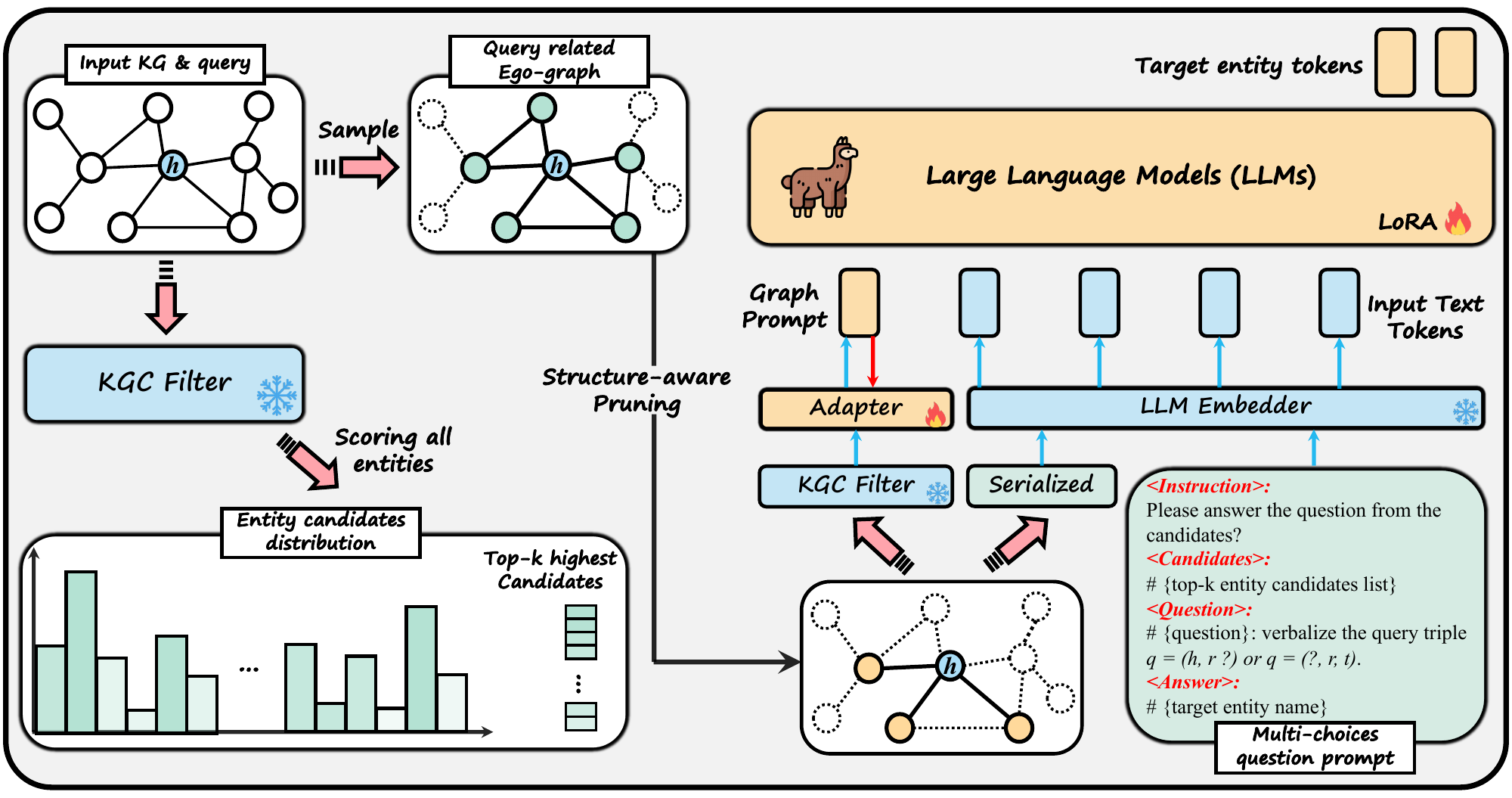}
    \caption{The overall framework of FtG. For a query triple, we employ a KGC filter to obtain top-$k$ candidates and construct corresponding multiple-choice question instruction. Then we sample the ego-graph of query entity and prune irrelevant neighbors. To bridge the gap between graph and text, we encode the pruned ego-graph into a soft graph token and map the graph token into text embedding space with a lightweight adapter. The target entity name is generated with soft graph token, textual serialization of pruned ego-graph, and instruction prompt.}
    \label{framework}
\end{figure*}

\section{Preliminaries}
\textbf{Knowledge Graph Completion (KGC).} Knowledge graph (KG) is commonly composed of a collection of fact triples. Let $\mathcal{G}=(\mathcal{E}, \mathcal{R}, \mathcal{T})$ be a KG instance, where $\mathcal{E}, \mathcal{R},$ and $\mathcal{T}$ represent the set of entities, relations, and triples, respectively. Each triple $(h, r, t) \in \mathcal{E} \times \mathcal{R} \times \mathcal{E}$ describes the fact that a relation $r$ exists between head entity $h$ and tail entity $t$. Given an incomplete triple $(h,r,?)$ or $(?,r,t)$ as query $q$, knowledge graph completion aims to predict the missing tail or head entity. In conventional KGC models, they learn specific structural embeddings for KGs, and the missing entity is predicted by finding the highest score $f(h,r,e)$ or $f(e,r,t), \forall e\in \mathcal{E}$, where $f$ is the model-specific scoring function.

\noindent \textbf{Instruction Tuning for KGC.} Instruction tuning refers to fine-tuning LLMs to follow human-curated instructions, enabling adaptation of LLMs to specific tasks. When applying LLMs to the KGC task, an instruction tuning sample comprises an instruction prompt and input-output pair. The instruction prompt $\mathcal{I}$ (e.g., "Predict the missing tail entity") is definition of KGC task for LLMs to comprehend and execute. The input $X$ is the verbalization of the query $q$ described in natural language. The instruction tuning process aims to strictly generate the missing entity in natural language given the instruction and the query input: $Y=\text{LLM}_{\theta}(\mathcal{I}, X)$, where $\theta$ are the parameters of LLM. The prevalent Negative Log-Likelihood Loss in language modeling is selected as training objective, which can be formed as:
\begin{equation}
    \mathcal{L}(\theta)=-\sum_{i=1}^{L}\log P_{\theta}(Y_i\vert\mathcal{I},X,Y_{<i}),
\end{equation}
where $Y_{<i}$ represents the prefix of missing entity name sequence $Y$ up to position $i-1$, $P_{\theta}(Y_i\vert\mathcal{I},X,Y_{<i})$ represents the probability of generating token $Y_i$, and $L$ is the sequence length of $Y$.

\section{Methodology}
In this section, we first provide an in-depth description of our \textit{filter-then-generate} paradigm in Sec.~\ref{ftg}, specifically designed for KGC task. Based on such paradigm, to bridge the gap between graph structure and text, we further introduce two novel modules: 1) a flexible ego-graph serialization prompt in Sec.~\ref{prompt}, which can effectively convey the structural information around the query triple, and 2) a structure-text prefix adapter in Sec.~\ref{adapter}, to map graph structure features into the text embedding space. Finally, we detail the instruction tuning strategy in Sec.~\ref{training}, focusing on efficient adaptation to KGC task. The overall architecture of our proposed model is illustrated in Figure~\ref{framework}.

\subsection{Filter-then-Generate Paradigm}\label{ftg}

To address the challenge of large entity candidate set, we propose a novel \textit{filter-then-generate} paradigm for LLMs on KGC task. As its name implies, we utilize a filter to eliminate unlikely entities and retain only the top-$k$ candidates. LLMs then generate the target entities conditioned on the query and candidates list. Within our paradigm, given a query $q=(h,r,?)$ or $q=(?,r,t)$, we employ a conventional structure-based KGC method as filter to score each entity $e$ in KG $\mathcal{G}$ and retain top-$k$ highest scoring entities as candidate set $C_k=[e_{1}, e_{2}, e_3 ..., e_{k}]$. The idea behind this paradigm is that conventional shallow embedding based models are good at easy samples but fail to discriminate target entities from a set of hard samples. Then we reformulate the KGC task into the form of multiple-choice question and design a simple instruction template that prompts LLMs to generate answer:
\begin{tcolorbox}
\textbf{Instruction:} Please answer the following question and select only one answer from the candidates that is most relevant to the question. \\
\textbf{Question:} <Verbalization of query $q$> \\
\textcolor{red}{\textbf{[Optional] Context:} <Ego-graph serialization prompts>}\\
\textbf{Candidates:} <Candidate set $C_k$> \\
\textbf{Answer:} <Target entity name>
\end{tcolorbox}

In above instruction template, we adopt the same verbalization as \citet{kg-llama} to transform the query triple into a simple question, and the context is optional, which conveys the structural information around the query described in following section. Essentially, our \textit{filter-then-generate} pipeline demonstrates following advantages. Firstly, it can rescue LLMs from a large number of candidate entities, enabling LLMs to leverage their own knowledge and reasoning abilities to identify the target entity from a group of candidate entities, which go beyond the abilities of conventional methods but could be well solved by LLMs. Additionally, the multiple-choice question format effectively avoids LLMs to output uncontrollable text.

\subsection{Ego-graph Serialization Prompt}\label{prompt}
In our paradigm, we aim to exploit semantic comprehension and reasoning capability of LLMs for KGC task. Nevertheless, transforming the query triple into a text-based prediction inevitably neglects the structural information of KG, which is an important feature for KGC task.Moreover, understanding graph structures using LLM continues to be a challenge, and although there has been some exploration into designing prompts to convey structural information, a comprehensive solution is still lacking.

To incorporate the structural information of KG into LLMs, we design an ego-graph serialization prompt. Instead of accessing to the entire KG, we extract the 1-hop ego-graph \citep{ego-graph} around the query entity, which characterizes the first order connectivity structure of entity. Considering that not all neighborhoods are useful for query, and some of them even introduce additional noise, we employ structure embeddings of KGs to sample more informative neighbors. Specifically, given a KG $\mathcal{G}$ and the query triple $q=(h,r,?)$ under the tail entity prediction setting (same in head entity prediction), we first sample the both incoming and outgoing triples of $h$ as the 1-hop ego-graph $\mathcal{N}_h=\{(h,r^{\prime},e)\in \mathcal{G}\} \cup \{(e, r^{\prime},h) \in \mathcal{G} \}$. Then let $\mathbf{E} \in \mathbb{R}^{\vert \mathcal{E} \vert \times d_s}$ and $\mathbf{R} \in \mathbb{R}^{\vert \mathcal{R} \vert \times d_s}$ denote the structural entity embedding matrix and relation embedding matrix, respectively, and $d_s$ is structural embedding dimension. The structure embedding matrices are provided by the KGC model adopted as the filter of FtG. We take in query $q=(h,r,?)$ and the 1-hop ego-graph $\mathcal{N}_h$ to extract the most relevant neighbors $\widetilde{\mathcal{N}_h}$ as follows:
\begin{equation}
\begin{aligned}
    \widetilde{\mathcal{N}_h} = \{ & (h', r', t') \mid (h', r', t') \in \mathcal{N}_h \\
    & \text{ and } \text{cos}(\boldsymbol{h'} \parallel \boldsymbol{r'}, \boldsymbol{h} \parallel \boldsymbol{r}) > \epsilon \}
\end{aligned}
\end{equation}
where $\boldsymbol{h}\in \mathbb{R}^{1\times d_s}$ and $\boldsymbol{r} \in \mathbb{R}^{1\times d_s}$ represent the structural embedding of $h$ and $r$, $\text{cos}(\cdot,\cdot)$ is the cosine similarity, $\epsilon$ is the threshold, and $\parallel$ is the concatenation operation. 

After obtaining the extracted ego-graph $\widetilde{\mathcal{N}_h}$, we follow existing work \citep{reasoninglm} perform breadth-first search (BFS) serialization to linearize it into a textual sentence. Specifically, starting from the entity $h$, we perform BFS to visit all entities in $\widetilde{\mathcal{N}_h}$. We then concatenate all the visited triples in the order of their traversal during the BFS process and remove duplicate entities, resulting in a long sequence, denotes as:
\begin{equation}
    S_{\widetilde{\mathcal{N}_h}}=\{h, r_1, e_1, r_2, e_2, \cdots, r_m, e_m\},
\end{equation}
where $m$ is the number of triples in $\widetilde{\mathcal{N}_h}$. 

\subsection{Structure-Text Adapter}\label{adapter}
\noindent \textbf{Graph Encoding and Adaption.} While our ego-graph serialization prompt has captured the local structure information around the query, the linearization process inevitably loses the connective pattern of the graph. Therefore, we propose a soft prompt strategy to couple the KG structure and text information in a contextualized way. Given the pruned ego-graph $\widetilde{\mathcal{N}_h}$, we obtain the ego-graph representation through \textit{parameter-free} message passing on encoded structure features, and map the graph representation into the embedding space of LLM via a trainable projection matrix $\mathbf{W}_p \in \mathbb{R}^{d_s \times d_x}$:
\begin{equation}
\mathbf{S}_{\widetilde{\mathcal{N}_h}}= \frac{1}{\vert \widetilde{\mathcal{N}_h} \vert}\sum_{e^\prime \in \widetilde{\mathcal{N}_h}} \boldsymbol{e^\prime}, \quad \mathbf{S}_{\widetilde{\mathcal{N}_h}}^\prime=\mathbf{W}_p\cdot \mathbf{S}_{\widetilde{\mathcal{N}_h}},
\end{equation}
where $\mathbf{S}_{\widetilde{\mathcal{N}_h}}^\prime$ is the projected ego-graph representation, $\boldsymbol{e^\prime} \in \mathbb{R}^{1\times d_s}$ is corresponding entity structural embedding, and $d_x$ denotes the dimension of embedding space of LLMs. We do not explore more complex adaptation schemes (e.g., cross-attention) because they require extra graph-text pairs for pre-training. Moreover, such straightforward linear projection allows us to iterate data-centric experiments quickly, which has been proven effective in visual-text alignment \citep{project}.

\noindent \textbf{Target Entity Generation.} Given a query $q=(h,r,?)$ and ego-graph serialization sequence $S_{\widetilde{\mathcal{N}_h}}$, we formulate them to corresponding textual version and obtain the input of the LLM, denoted as $X=X_q + X_{S_{\widetilde{\mathcal{N}_h}}}$. Let $\mathbf{X} \in \mathbb{R}^{\vert X\vert \times d_x}$ denote the textual content embeddings of input, where $\vert X \vert$ is the token length of $X$. We  concatenate the soft graph token and input embeddings as final input of LLMs, i.e., $\mathbf{X}^\prime =  \mathbf{S}_{\widetilde{\mathcal{N}_h}}^\prime \parallel \mathbf{X}$. In this way, the structure information can interact frequently with the textual information, enabling LLMs to leverage the underlying graph structure. Finally, our optimization objective is to maximize the probability of generating the target entity name $Y_t$ by maximizing:
\begin{equation}
    P(\mathbf{Y}_t\vert \mathbf{X}^{\prime}, \mathbf{X}_{\mathcal{I}})=\prod_{i=1}^{L}P_\theta(y_i\vert \mathbf{S}_{\widetilde{\mathcal{N}_h}}^\prime \parallel \mathbf{X}, \mathbf{X}_{\mathcal{I}}, \mathbf{Y}_{t,<i}),
\end{equation}
where $\mathbf{X}_{\mathcal{I}}$ denotes the representation of instruction tokens, and $L$ is the token length of target entity.

\begin{table*}[t]
 \centering
 \resizebox{\linewidth}{!}{
 \begin{tabular}{lcccccccccccc}\toprule
    \multirow{2}{*}{\textbf{Model}} & \multicolumn{4}{c}{\textbf{FB15k-237}} & \multicolumn{4}{c}{\textbf{CoDEx-M}} & \multicolumn{4}{c}{\textbf{NELL-995}}
    \\\cmidrule(lr){2-5}\cmidrule(lr){6-9}\cmidrule(lr){10-13}
             & MRR & H@1 & H@3 & H@10  & MRR & H@1 & H@3 & H@10  & MRR & H@1 & H@3 & H@10 \\\midrule \specialrule{0em}{1.5pt}{1.5pt}
    \textbf{\textit{Structure-Based Methods}} \\
    TransE~\citep{transe} & .279 & .198 & .376 & .441 & .303 & .223 & .336 & .454 & .401 & .344 & .472 & .501 \\
    DistMult~\citep{distmult} & .281 & .199 & .301 & .446 & .223 & .145 & .245 & .383 & .485 & .401& .524 & .610 \\
    ComplEx~\citep{complex} & .278 & .194 & .297 & .450 & \underline{.337} & \underline{.262} & \underline{.370} & \textbf{.476} & .482 & .399 & .528 & .606 \\
    ConvE~\citep{conve} & .312 & .225 & .341 & .497 & .318 & .239 & .355 & .464 & .491 & .403 & .531 & .613 \\
    RotatE~\citep{rotate} & .338 & .241 & .375 & .533 & .302 & .219 & .341 & .461 & .483 & .435 & .514 & .565 \\
    KG-Mixup~\citep{KG-mixup} & \underline{.358} & .264 & - & \underline{.548} & .319 & .242 & - & .465 & \underline{.522} & \underline{.458} & - & \underline{.621} 
    \\\midrule
    \textbf{\textit{PLM-Based Methods}} \\
    
    GenKGC~\citep{genkgc} & - & .192 & .355 & .439 & - & - & - & - & - & - & - & - \\
    KG-S2S~\citep{kgs2s} & .336 & .257 & .373 & .498 & .246 & .186 & .268 & .372 & .392 & .324 & .438 & .511 \\
    CSProm-KG~\citep{csform} & \underline{.358} & \underline{.269} & \underline{.393} & .538 & .320 & .243 & .355 & .464 & .508 & .438 & .548 & \textbf{.626} \\
    \midrule
    ChatGPT~\citep{chatgpt1shot} $\clubsuit$ & - & .097 & .112 & .124 & - & - & - & - & - & - & - & - \\
    PaLM2-540B~\citep{Palm2} $\clubsuit$ & - & .115 & .166 & .217 & - & - & - & - & - & - & - & - \\
    KG-LLaMA-7B~\citep{kg-llama} $\diamondsuit$ & .238 & .165 & .272 & .423 & .179 & .159 & .200 & .204 & .397 & .388 & .405 & .406 
    \\\midrule
    \textbf{FtG}(Ours) & \textbf{.392} & \textbf{.321} & \textbf{.413} & \textbf{.542}  & \textbf{.395} & 
    \textbf{.352} & \textbf{.407} & \underline{.473} & \textbf{.538} & \textbf{.479} & \textbf{.573} & \textbf{.626}
    \\\bottomrule
 \end{tabular}}
 \caption{The performance of FtG and baselines on three KG datasets. $\clubsuit$ denotes results are obtained through evaluating all test triples from the \citet{distill}. $\diamondsuit$ means that partial unreported results are obtained through the implementations as \citet{kg-llama}.}
 \label{main_result}
\end{table*}

\noindent \textbf{Connection to Graph Neural Networks.} Our model shares essential mechanism similarities with existing GNNs, thus covering their advantages. First, due to the input length limitation of LLMs, our ego-graph serialization prompt for the query entity is aligned with GraphSAGE \citep{graphsage}. And our similarity-based extraction module resembles graph regularization techniques like DropEdge \citep{dropedge}. Additionally, our structure-text adapter carries structure features that can interact with text semantic features deeply in the encoding phase. Causal attention in LLMs can be regarded as an advanced weighted average aggregation mechanism of GAT \cite{gat}, facilitating our model to effectively model the varying importance of different neighbors to the central entity. Therefore, our framework integrates inductive bias required for graph tasks and enhances the graph structure understanding capability of LLMs. 

\subsection{KGC-Specific Instruction Tuning Strategy}\label{training}
The instruction tuning process aims to customize the  reasoning behavior of LLM to meet the specific constraints and requirements of KGC task. An example of our instruction data can be seen in Appendix~\ref{our_prompt}. During the training process, we always keep the parameters of KGC filter frozen, and update both the weights of the projection layer and LLM. Considering the computation overhead of full-parameters updates for LLM, we employ low-rank adaptation (i.e., LoRA \citep{lora}) due to its simple implementation and promising performances \citep{few}. This approach freezes the pre-trained model parameters and updates the parameters of additional trainable weight matrix $\mathbf{W}\in \mathbb{R}^{d_1\times d_2}$ by decomposing it into a product of two low-rank matrices: $\mathbf{W}=\mathbf{B}\mathbf{A}$, where $\mathbf{B} \in \mathbb{R}^{d_1\times r}$, $\mathbf{A} \in \mathbb{R}^{d_2\times r}$, and $r \ll \text{min}(d_1,d_2)$. Hence, LoRA can effectively adapt the LLM to KGC task while requiring little memory overhead for storing gradients.

\section{Experiment}
\subsection{Experimental Setup} 
We employ three widely-used KG datasets for our evaluation: FB15k-237 \citep{fb15k-237}, CoDEx-M \citep{codex}, and NELL-995 \citep{nell}. Detailed dataset statistics are shown in Appendix~\ref{data}. And the baselines adopted in our experiments are shown in Appendix~\ref{baseline}. In our implementation, we use LLaMA2-7B \citep{llama2} as the LLM backbone. We employ RotatE \citep{rotate} as our filter for its simplicity and lightweight nature. The effect of different filters is discussed in Sec.~\ref{discuss}. We report Mean Reciprocal Rank (MRR) and Hits@N (N=1,3,10) metric following the previous methods. Specific implementation details please refer to Appendix~\ref{implementation}.

\subsection{Main Results}
Table~\ref{main_result} displays the results of our experiments. Overall, we can observe that FtG achieves consistent and significant improvement on both datasets across  most metrics, which demonstrates the effectiveness of our proposed FtG.

Compared to structure-based baselines, FtG showcases remarkable performance, which demonstrates the capability of FtG to understand and leverage graph structure. Additionally, compared to RotatE, which is employed as the filter in our model, FtG achieves Hits@1 improvements of  $33.2\%$, $60.7\%$, and $10.11\%$ across three datasets, indicating that \textit{filter-then-generate} paradigm can effectively incorporate the strength of RotatE and LLMs, enabling FtG to leverage knowledge memorization and reasoning ability of LLM to address indistinguishable entity candidates. Compared to the sparse NELL-995, FtG improves more in the remaining two datasets, suggesting that FtG can fully utilize the structural information of the KG.

For the PLM-based baselines, FtG outperforms the SOTA method CSProm-KG by a substantial margin, indicating the superiority of our method. Focusing on the LLMs-based methods, we can find that even with instruction fine-tuning on KGs, LLMs still yield inferior performance. The reason is that directly eliciting LLMs to generate answers is prone to be influenced by hallucination of LLMs, leading to uncontrollable responses. In comparison, FtG achieves substantial improvements across three datasets, indicating that FtG can effectively harness and unleash the capability of LLMs. 

\subsection{Ablation Study}
In this subsection, we conduct an ablation study to investigate the individual contributions of different components in FtG. The results and meanings of various variants are reported in Table~\ref{ablation}. The results reveal that all modules are essential because their absence has a detrimental effect on performance.

\begin{table}[ht]
    \centering
    \resizebox{\linewidth}{!}{
    \begin{tabular}{ccc|cccc}\toprule
    \multirow{2}{*}{FGP} & \multirow{2}{*}{ESP} & \multirow{2}{*}{STA} & \multicolumn{2}{c}{\textbf{FB15k-237}} & \multicolumn{2}{c}{\textbf{NELL995}} \\\cmidrule(lr){4-5}\cmidrule(lr){6-7}
     & & & MRR & H@1 & MRR & H@1 \\\midrule
     \CheckmarkBold & \CheckmarkBold & \CheckmarkBold & \cellcolor{mygray}\textbf{.392} & \cellcolor{mygray}\textbf{.321} & \cellcolor{mygray}\textbf{.538} & \cellcolor{mygray}\textbf{.479}\\\midrule
      \CheckmarkBold & \XSolidBrush & \XSolidBrush & $.363_{(\downarrow .029)}$ & $.279_{(\downarrow .042)}$ & $.534_{(\downarrow .004)}$ & $.469_{(\downarrow.010)}$ \\
       \CheckmarkBold & \CheckmarkBold & \XSolidBrush & $.374_{(\downarrow .018})$ & $.295_{(\downarrow.026)}$ & $.535_{(\downarrow .003)}$ & $.471_{(\downarrow .008})$ \\
     \CheckmarkBold & \XSolidBrush & \CheckmarkBold & $.382_{(\downarrow .010)}$ & $.306_{(\downarrow .015)}$ & $.532_{(\downarrow .006)}$ & $.472_{(\downarrow .007)}$ \\
      \XSolidBrush & \XSolidBrush & \XSolidBrush & $.238_{(\downarrow .154)}$ & $.165_{(\downarrow.156)}$ & $.397_{(\downarrow .141)}$ & $.388_{(\downarrow .091)}$ \\
      \bottomrule
    \end{tabular}}
    \caption{Ablation for the FtG in FB15k-237 and NELL995. FGP denotes \textit{filter-then-generate} paradigm. ESP denotes ego-graph serialization prompt. STA denotes structure-to-text adapter.}
    \label{ablation}
\end{table}

Specifically, to demonstrate the effectiveness of \textit{filter-then-generate} paradigm, we directly adopt multiple-choice question instruction to fine-tune LLaMA-7b, and the results are shown in Table~\ref{ablation} Line 2. We observe that our paradigm significantly outperforms the base model that directly adopts instruction tuning (the last line). This demonstrates filter-then-generate paradigm can harness and unleash the capability of LLMs. Moreover, by comparing the variant with ego-graph serialization prompt (Line 3) and Line 2, we find that neighborhood information surrounding the query entity can facilitate LLMs to perform KGC task. And the effect of ego-graph serialization prompt is more significant in FB15k-237, which is due to the fact that NELL-995 is a more sparse KG (the average node degree is lower than 2).

Besides, comparing the Line 2 and Line 4, we can find that our proposed graph soft prompt can achieve impressive Hits@1 improvements of $9.7\%$ on FB15k-237, which demonstrates the soft graph prompt with lightweight adapter can deeply encode the inherent structural characteristics of the KG.

\subsection{Discussion}\label{discuss}
In this section, we conduct a multifaceted performance analysis of FtG by answering the following questions. More analysis please refer to Appendix~\ref{ins}.


\begin{figure}[ht]
    \subfigure[RotatE on FB15k-237.]{
    \centering
    \includegraphics[width=0.49\linewidth]{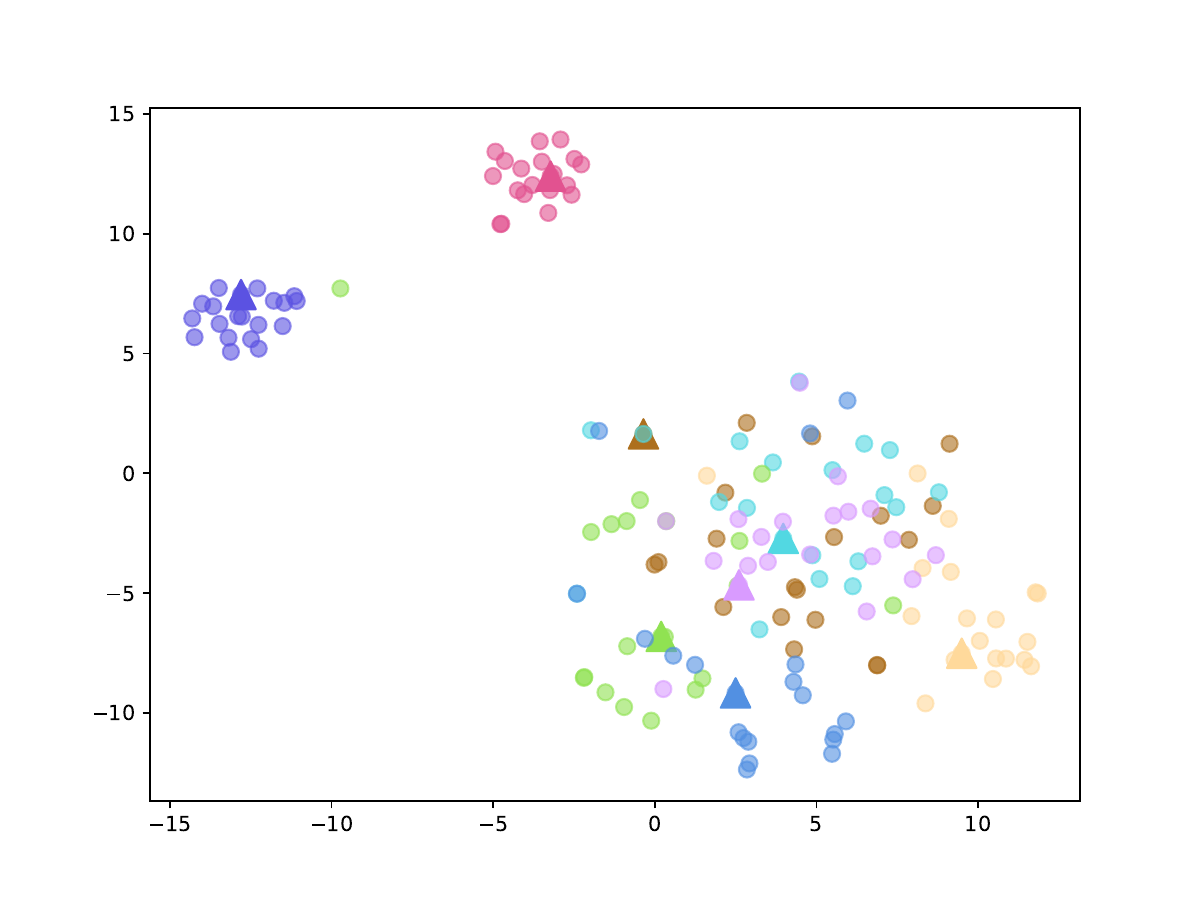}
    }\subfigure[FtG on FB15k-237.]{
    \centering
    \includegraphics[width=0.49\linewidth]{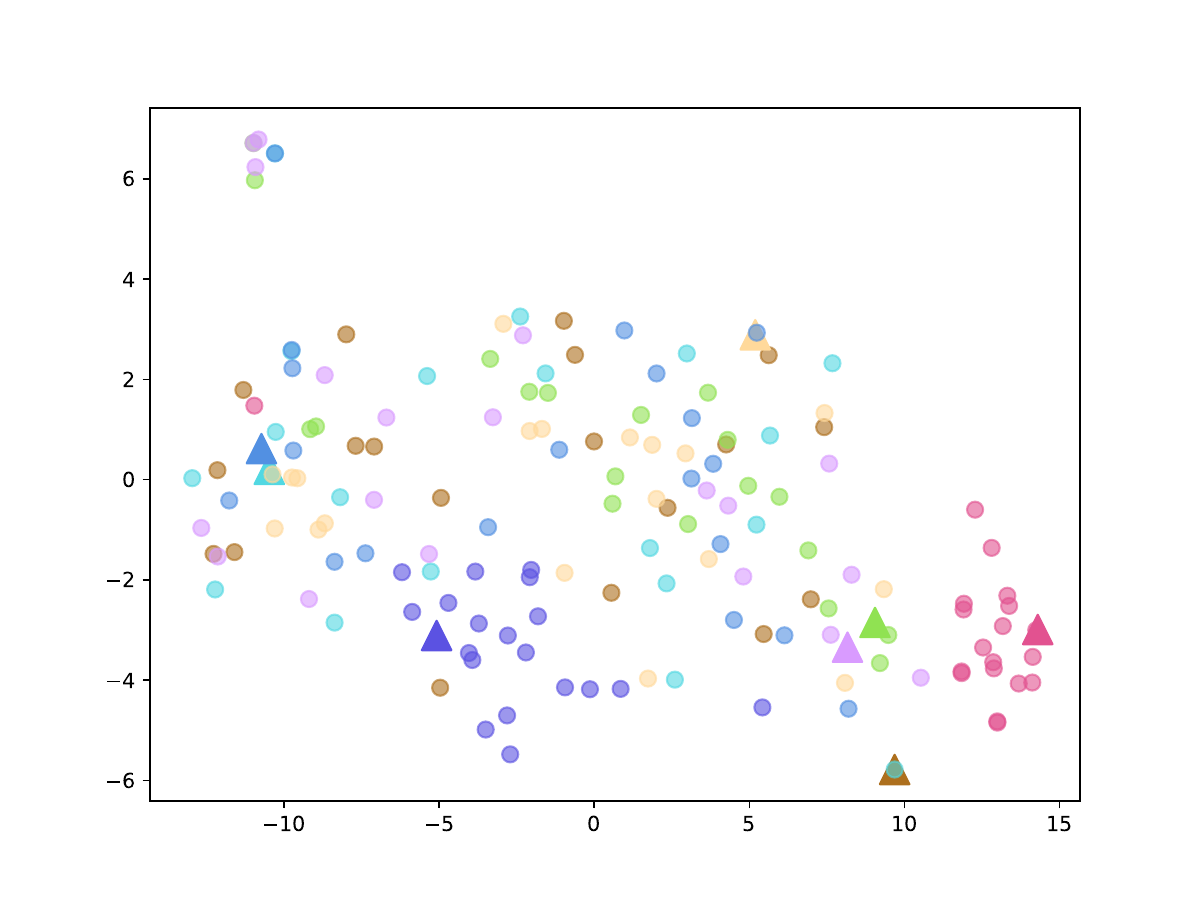}
    }

    \subfigure[RotatE on CoDEx-M.]{
    \centering
    \includegraphics[width=0.49\linewidth]{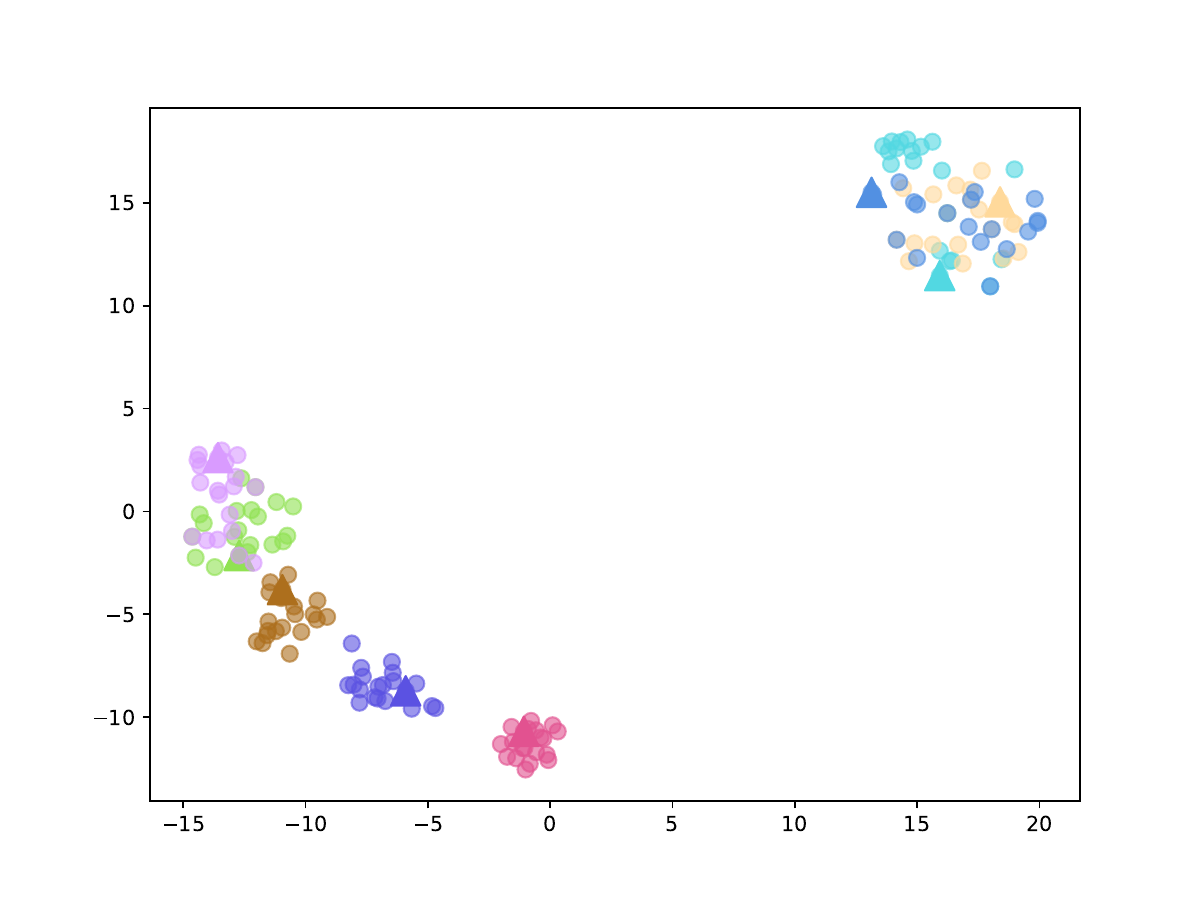}
    }\subfigure[FtG on CoDEx-M.]{
    \centering
    \includegraphics[width=0.49\linewidth]{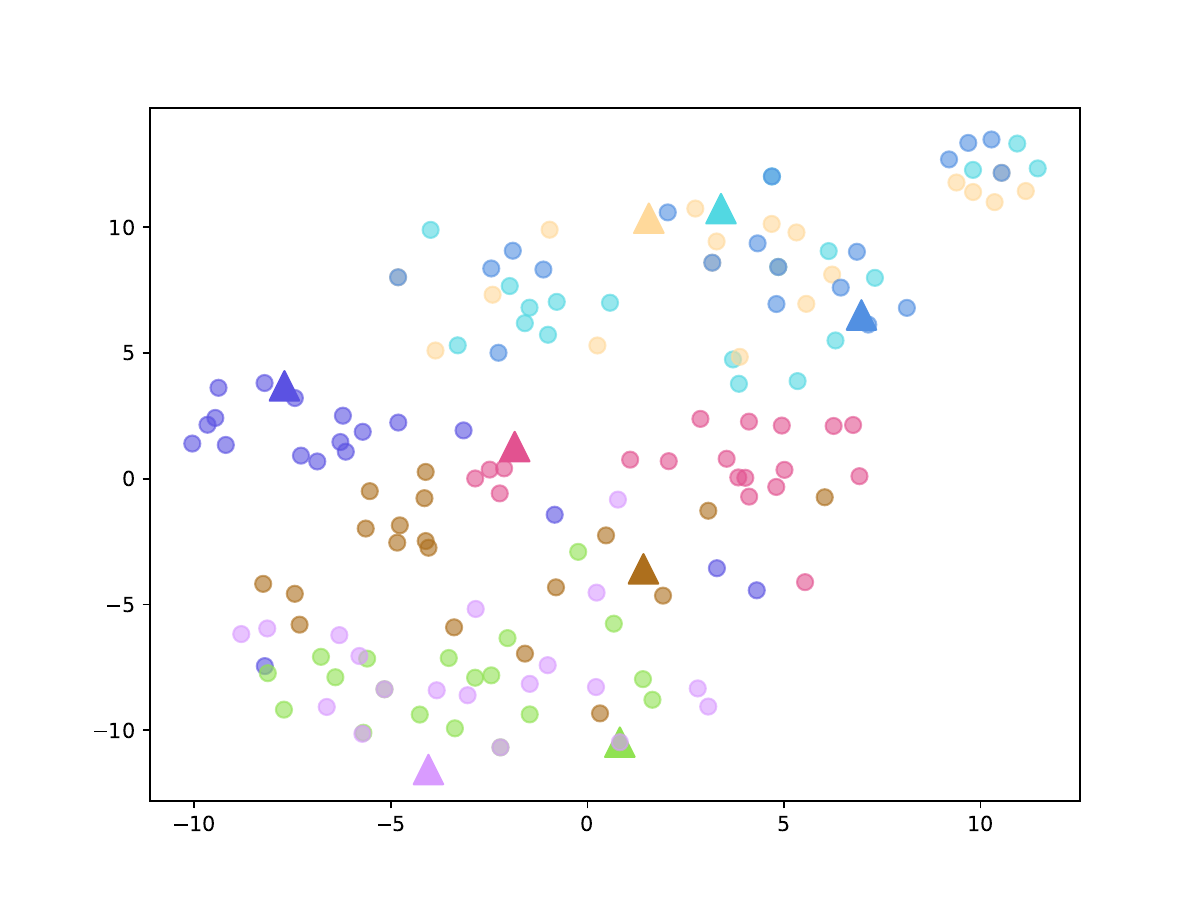}
    }

    \subfigure[RotatE on NELL-995.]{
    \centering
    \includegraphics[width=0.49\linewidth]{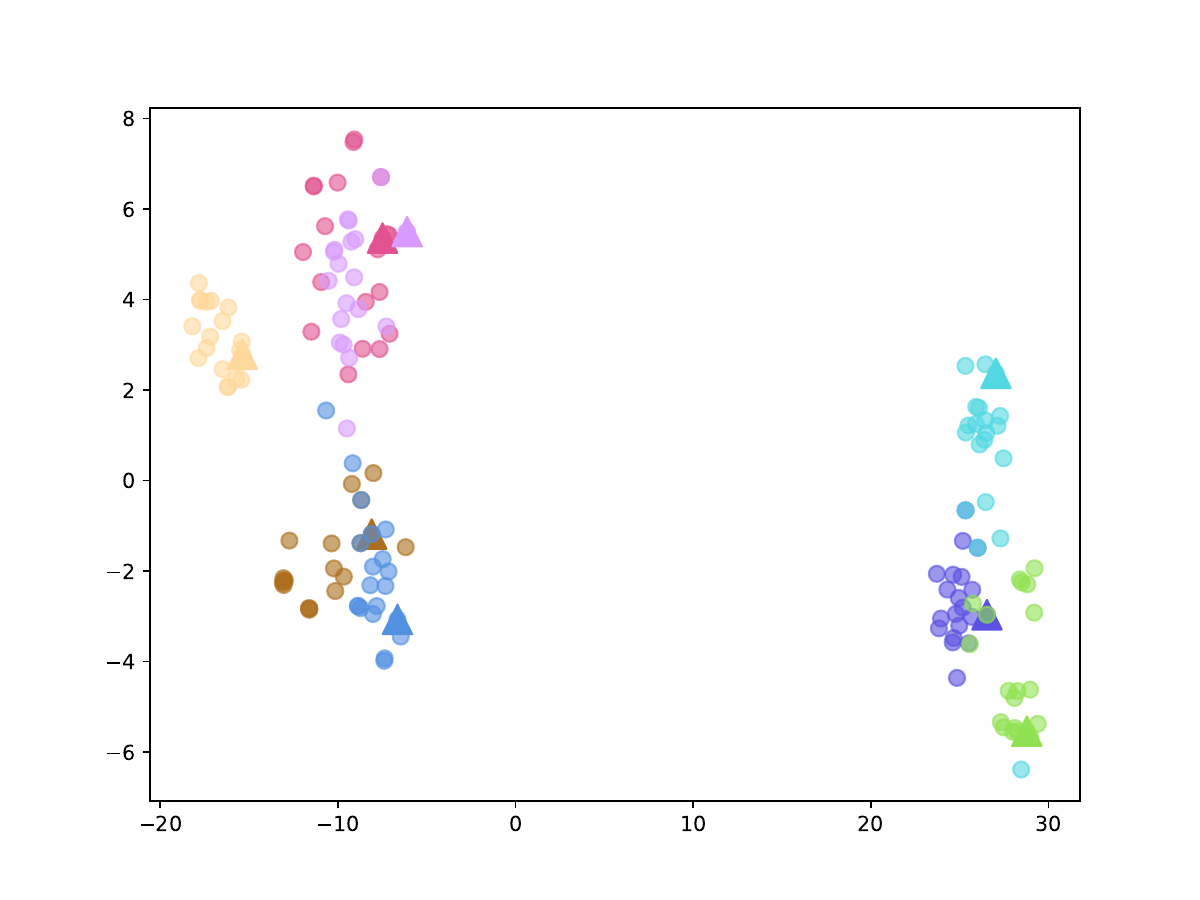}
    }\subfigure[FtG on NELL-995.]{
    \centering
    \includegraphics[width=0.49\linewidth]{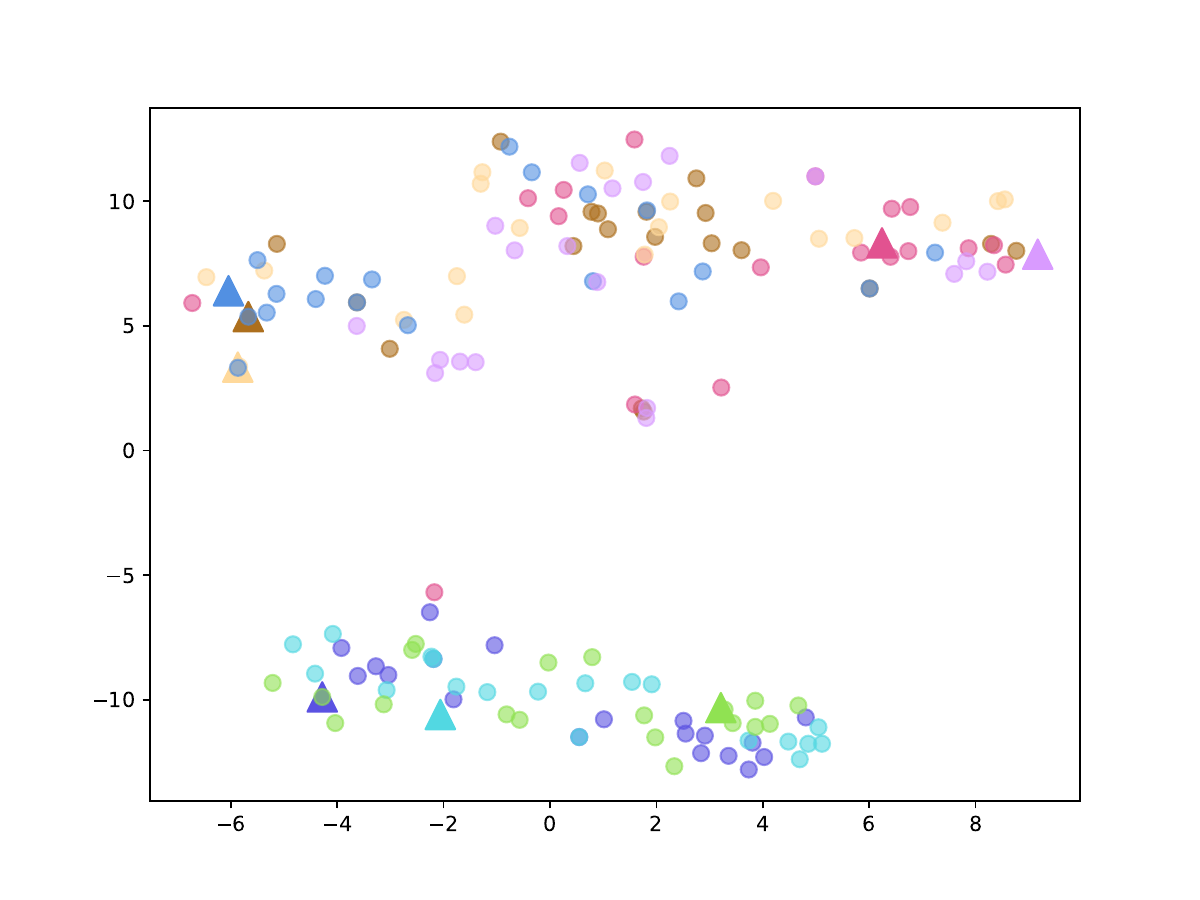}
    }
    \caption{Visualization of candidate entity embeddings in CoDEx-M and NELL-995. Each color denotes a query, and candidate entities of the same color belong to the same query. $\triangle$ denotes the target entity.}
    \label{more_vis}
\end{figure}

\noindent \textbf{Q1: Why can FtG make impressive gains?} Our motivation for proposing FtG is that we argue that \textbf{\textit{LLMs are not good KGC reasoner, but strong discriminator of hard samples.}} We select several test samples where RotatE prediction failed in FB15k-237, and visualize the top-20 highest scoring entities embeddings of each query in Figure~\ref{more_vis} (a). Dots of the same color represent entities that are candidates under the same query, and $\triangle$ is the target entity. Obviously, we can find that these candidate entities are indistinguishable for RotatE, and they overlap in the embedding space. As shown in Table~\ref{hard}, these hard samples require additional contextual knowledge to be distinguished. In contrast, we visualize the candidate entity's hidden states in the last transformer-layer of FtG in Figure~\ref{more_vis} (b). The figure demonstrates that our FtG can well distinguish the target entity from the hard samples, which is attributed to the inherent contextual knowledge of LLMs. Besides, we observe similar visualization results on the CoDEx-M and NELL-995 datasets.

\begin{figure}[t]
    \centering
    \includegraphics[width=\linewidth]{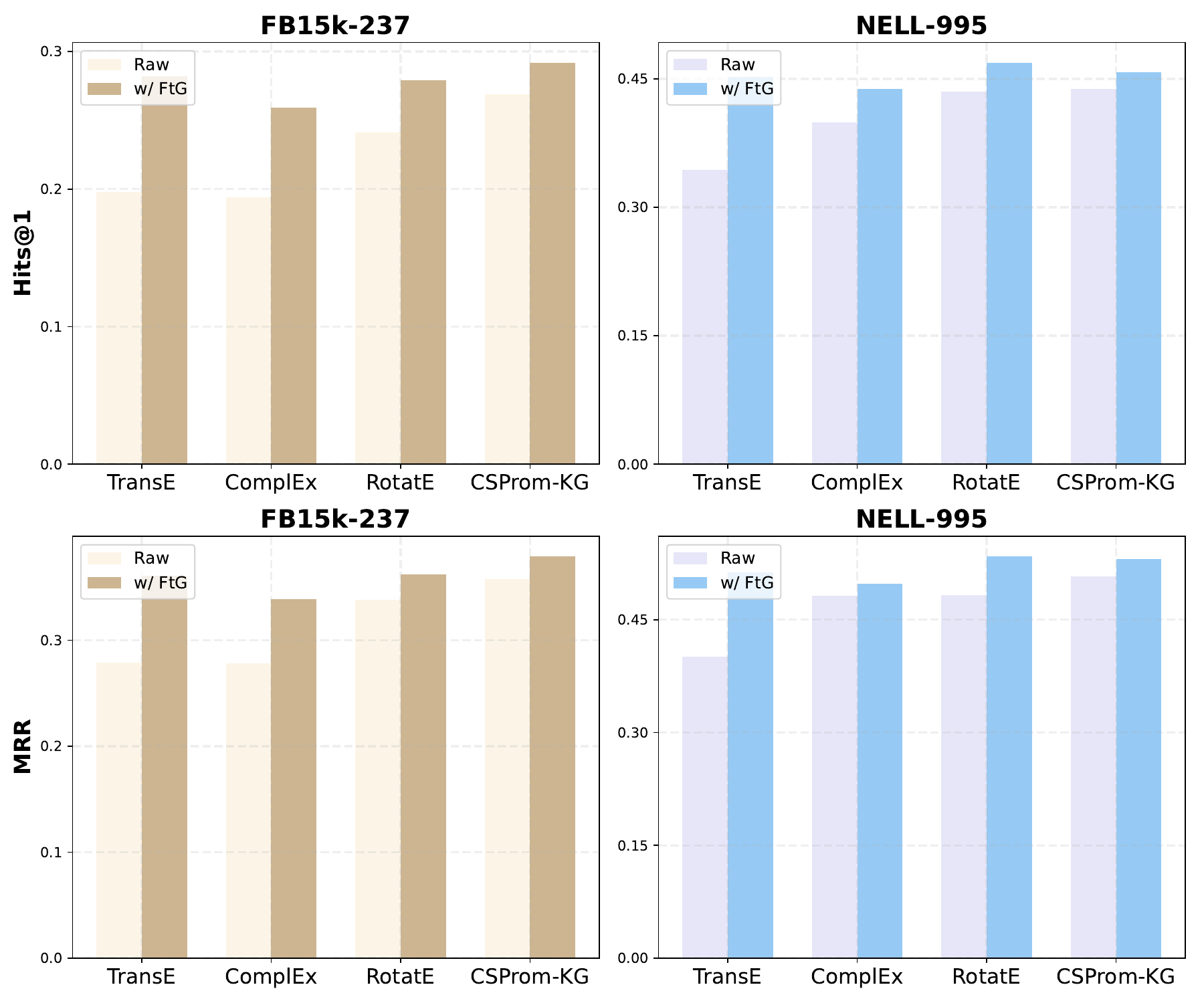}
    \caption{Performance (Hits@1\&MRR) of FtG with various KGC filter on FB15k-237 and NELL-995 dataset.}
    \label{various}
\end{figure}

\noindent \textbf{Q2: Is FtG compatible with existing KGC methods?} Here, we further evaluate whether our proposed FtG is robust enough when equipped with various KGC methods as a filter. Taking on a more challenging setting, we do not resort to retrain with different KGC methods. Instead, we load the trained LoRA weights directly and then switch different KGC filters for evaluation. Specifically, we employ a range of prevalent KGC methods, including structure-based TransE, ComplEx, RotatE, as well as PLM-based method CSProm-KG. Our results, as shown in Figure~\ref{various}, demonstrate that existing KGC methods achieve significant improvements when integrated with FtG, across both FB15k-237 and NELL-995 datasets. This suggests that FtG can effectively incorporate the strength of conventional KGC methods and LLMs, enabling FtG to leverage the reasoning ability of LLMs to address indistinguishable candidates. Furthermore, this also underscores our method's capacity to \textbf{enhance existing KGC methods in a plug-and-play manner}, demonstrating the versatility and effectiveness of FtG.

\noindent \textbf{Q3: Is structure-aware pruning necessary in ego-graph serialization prompt?} We replace the ego-graph serialization prompt in FtG with three other common heuristics,including random walk of the query entity, the entire one-hop ego-graph, and two-hop ego-graph. Empirical results are shown in Table~\ref{ego-graph}, their final prediction are outperformed by FtG. Notably, comparing the results of the entire one-hop ego-graph (Line 3) with FtG, we can see that structure-aware pruning plays a crucial role, especially on the FB15k-237 dataset, which is rich in graph structure information (average degree of entity is 37.4). These results demonstrate that our structure-aware pruning strategy can retain only the relevant information while filtering out the irrelevant neighbors. And it shares essential mechanism similarities with GAT \cite{gat}, thus covering the advantage of GAT. 
\begin{table}[ht]
    \centering
    \resizebox{\linewidth}{!}{
    \begin{tabular}{c|cccccc}\toprule
    \multirow{2}{*}{\textbf{Heuristics}} & \multicolumn{3}{c}{\textbf{FB15k-237}} & \multicolumn{3}{c}{\textbf{NELL995}} \\\cmidrule(lr){2-4}\cmidrule(lr){5-7}
      & MRR & H@1 & H@3 & MRR & H@1 & H@3 \\\midrule
     \rowcolor{mygray}\textbf{Structure-aware pruning} & \textbf{.392} & \textbf{.321} & \textbf{.413} & \textbf{.538} & \textbf{.479} & \textbf{.573}\\\midrule
      Random Walk & .356 & .286 & .402 & .531 & .468 & .567  \\
      Entire 1-hop Ego-graph & .368 & .313 & .405 & .536 & .476 & .568 \\
      2-hop Ego-graph & .382 & .308 & .404 & .528 & .463 & .565  \\
      \bottomrule
    \end{tabular}}
    \caption{Comparison of prediction performance with different heuristics.}
    \label{ego-graph}
\end{table}

\begin{figure}[ht]
    \centering
    \includegraphics[width=\linewidth]{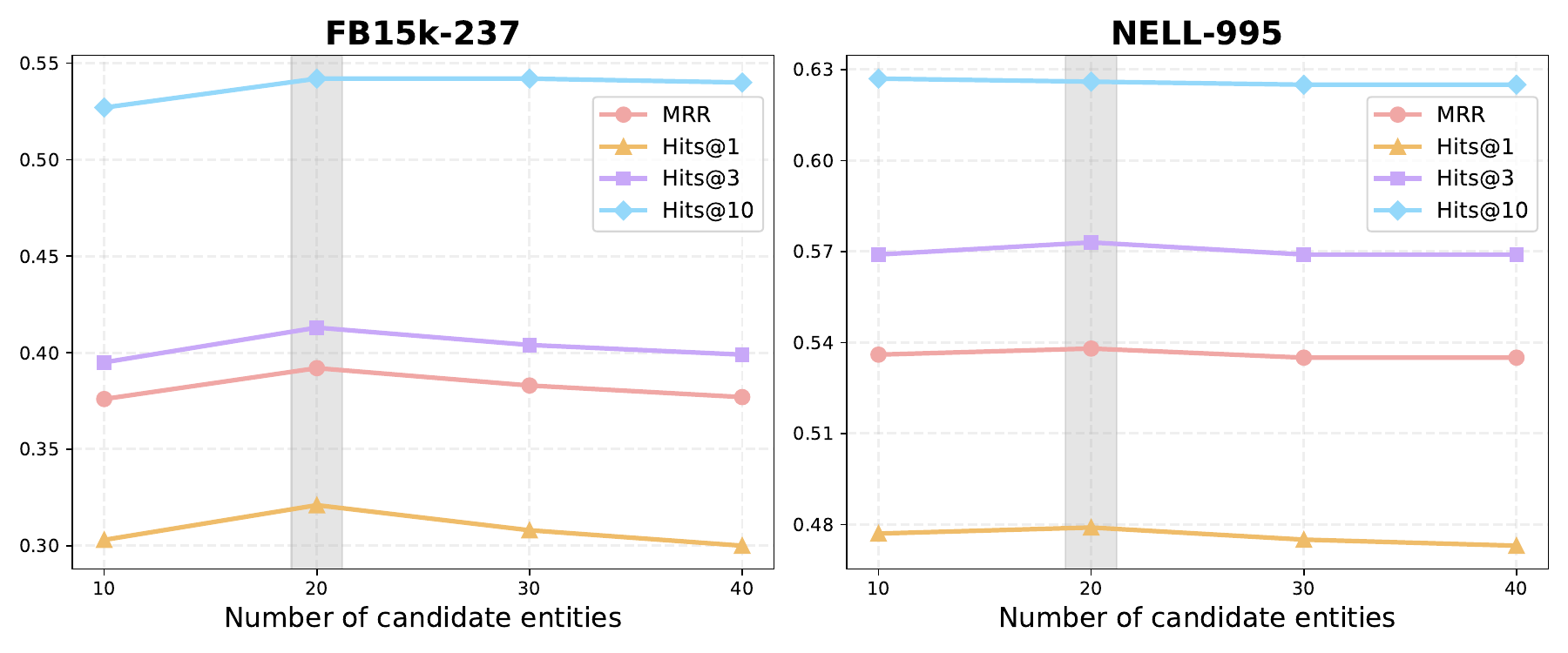}
    \caption{Performance of FtG with different sizes of candidate entities.}
    \label{can}
\end{figure}

\noindent \textbf{Q4: How does the number of candidate entities retained in the filtering stage affect the performance?} Here, we analyze the connection between the size of  candidate set and performance. Our results, as presented in Figure~\ref{can}. We observe that when increasing the size of the candidate set, the variation in model performance is not significant. The impact of this hyperparameter on results is akin to the trade-off between accuracy and recall. A larger candidate set implies a higher likelihood of containing the correct entity but also means that the LLM needs to comprehend more entities. In this paper, We prefer to enable LLMs to focus on hard samples that conventional KGC can not solve them well with limited model capacity and data amount. Therefore, we finally chose to set the size of the candidate entities to 20.

\noindent \textbf{Q5: What are effects of \textit{filter-then-generate} paradigm?} We devise several variants to fully analyze the impact of the FtG paradigm LLMs. Specifically, these variants are:
\begin{itemize}
    \item LLaMA2-7B: directly prompt the LLM for KGC , the prompt format can refer to Table~\ref{prompt1}.
    \item LLaMA2-7B-FtG: we do not fine-tune the LLM and only adopt the FtG prompt.
    \item ChatGPT: we utilize the same prompt (refer to Figure~\ref{prompt_llm}) as \citet{distill} to evaluate.
    \item KG-LLaMA2-7B: adopt instruction-tuning to adapt LLaMA2-7B for KGC.
    \item $\text{FtG}*$: we only adopt \textit{filter-then-generate} paradigm to fine-tune the LLM, it can be regarded as a ablation vertion.
\end{itemize}

\begin{figure}[ht]
    \centering
    \includegraphics[width=\linewidth]{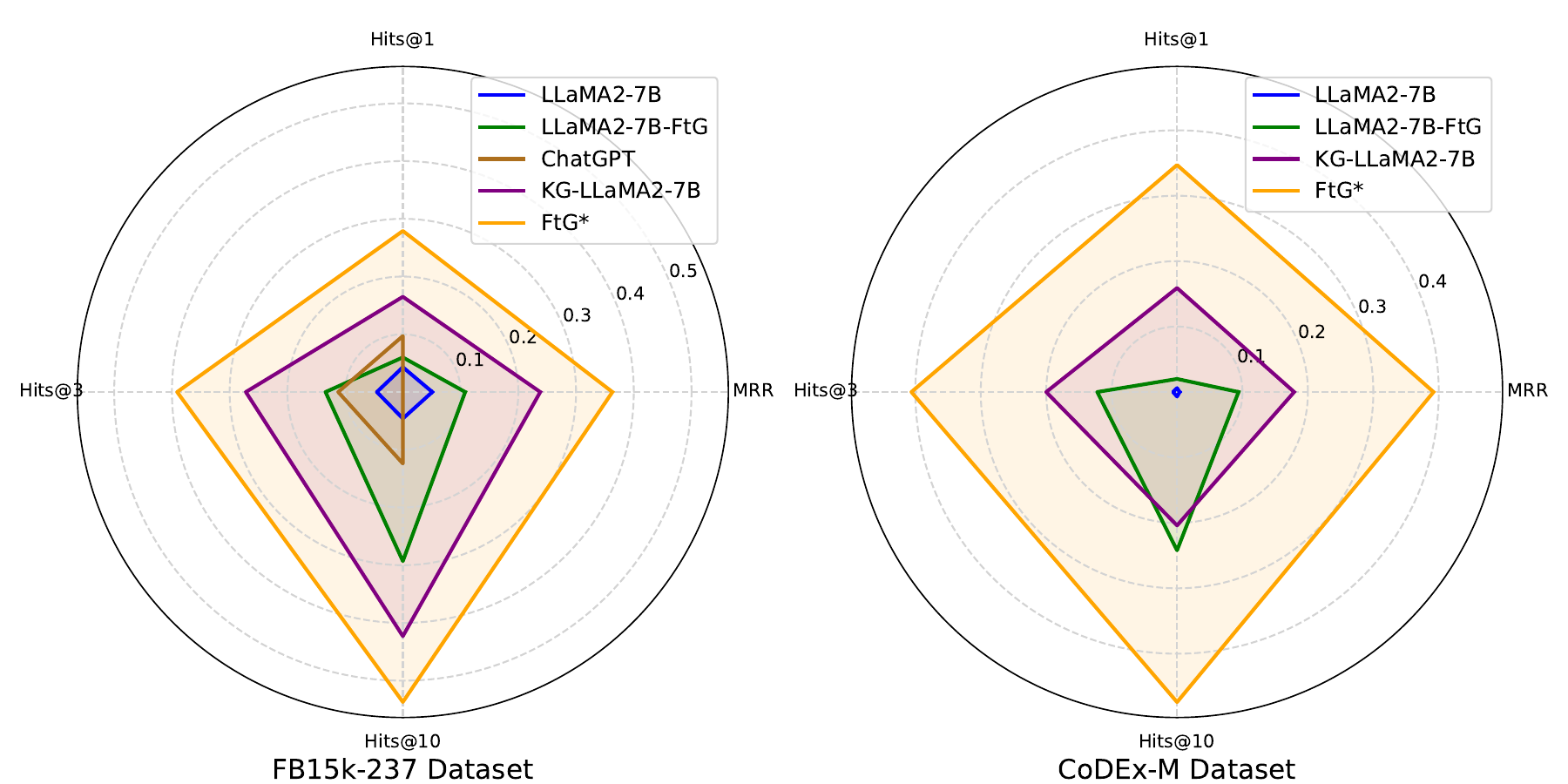}
    \caption{Effect of \textit{filter-then-generate} paradigm. The MRR metric of ChatGPT in FB15k-237 is unreported, to include it in our figure without causing distraction, we've set its value to 0 by default.}
    \label{radar}
\end{figure}

The results, as illustrated in Figure~\ref{radar}, indicate that the \textit{filter-then-generate} paradigm serves as an effective strategy for leveraging the capabilities of LLM in KGC. Especially, LLaMA2-7B with multiple-choice question prompt can outperform ChatGPT across some metrics in FB15k-237 dataset.



\section{Related Works}
\textbf{Structure-based KGC methods.} Early methods typically define a score function to evaluate the scores of triples through spatial measurement or latent matching. TransE \citep{transe} defines each relation as translation from the head entity to tail entity. RotatE \citep{rotate} further extends this idea in a complex space, enabling to model the symmetry relation pattern. Semantic matching methods, such as DistMult \citep{distmult} and ComplEx \citep{complex}, , leverage the semantic similarity to capture complex interactions among entities and relations. Additionally, RGCN \citep{rgcn}, CompGCN \citep{compgcn}, and SMiLE \citep{SMile} employ graph neural networks to model the graph structure patterns inherent in KGs. KG-Mixup \citep{KG-mixup} address the degree bias in KG, achieving promising results. 

\noindent\textbf{PLM-based KGC methods.}  KG-BERT \citep{kgbert} and StAR \citep{star} utilize cross-entropy object and fine-tune PLM to produce entity embeddings. SimKGC \cite{simkgc} convert the KGC task into a semantic matching task and introduce contrastive learning to model fine-grained semantics. However, these methods suffer from unstable negative sampling. KGT5 \citep{kgt5}, KG-S2S \citep{kgs2s}, and UniLP \citep{UniLP} further exploit T5 \citep{T5} with soft prompt to improve performance of generative KGC. \citet{fan} propose a retrieve-based method to probe knowledge from PLM for open KG completion. CSProm-KG \citep{csform} integrates PLM with a structure-based method to bridge the structure and text information, achieving SOTA performance.

Moreover, Large Language Models (LLMs) have revolutionized various tasks in natural language processing. Focusing on KGC task, \citet{chatgpt1shot} construct prompts and evaluate the performance of ChatGPT with in-context learning on KGC task. ChatRule \citep{chatrule} leverages ChatGPT to mine logical rules in KGs and applies these rules to make predictions. KICGPT \cite{kicgpt} is the first work to enhance KGC by using in-context learning to re-rank results and integrating LLMs with traditional models. With the development of techniques of LLMs in the open source community, KG-LLaMA \citep{kg-llama} makes the first step by applying instruction tuning to adapt LLaMA on KGC task. Additionally, some methods \citep{distill} distill contextual knowledge from LLMs to improve the quality of entity texts, thus benefiting existing PLM-based approaches. Nevertheless, existing approaches that utilize LLMs for KGC task have not demonstrated satisfactory performance, and it remains a challenge to apply LLMs in the KGC task. The detailed comparison between our FtG and existing LLM-based methods refer to Appendix~\ref{compare_llm}.

\section{Conclusions}
In this paper, we propose FtG, a instruction-tuning based method to enhance the performance of LLMs in KGC task. Our proposed \textit{filter-then-generate} paradigm can effectively harness the capabilities of LLMs. To further incorporate the structural information into LLMs, we devise an ego-graph prompt and introduce a structure-text adapter. Extensive experiments demonstrate the effectiveness of FtG. In the future, we plan to adapt our method to other relevant downstream tasks, such as recommendation and open question answering.

\section*{Limitations}
FtG can effectively harness the reasoning ability of LLMs and successfully incorporate the graph structural information into the LLMs, achieving substantial performance improvement on KGC task. However, the extremely large number of parameters in LLMs makes fine-tuning them resource-intensive. At the same time, LLMs are notoriously slow at decoding during inference. In our experiment, we use DeepSpeed \citep{deepspeed} to accelerate training and inference, but FtG remain slower than traditional methods due to its inherent scale. Besides, if the KGC filter is not able to recall the target entity within the top-$k$ candidates, FtG cannot make correct prediction. Therefore, a potential way to improve the effectiveness of FtG is to improve the success rate of target entity recall, and our FtG is more of a general framework to adapt LLM to KGC task.

\section*{Acknowledgements}
We would like to thank all the anonymous reviewers and area chairs for their comments. This research is supported by National Natural Science Foundation of China (U23A20316), General Program of Natural Science Foundation of China (NSFC) (Grant No.62072346), and funded by Joint\&Laboratory on Credit Technology.






\bibliography{anthology,custom}
\bibliographystyle{acl_natbib}

\appendix

\section{Dataset} \label{data}
\begin{table*}[t]
    \centering
    \resizebox{\linewidth}{!}{
    \begin{tabular}{c|c}\toprule
        Triple & (War on Terrorism, military/military conflict/combatants./military/military combatant group/combatants, Canada) \\\midrule
        Tail Prompt & War on Terrorism, military/military conflict/combatants./military/military combatant group/combatants? \\\midrule
        Head Prompt & What/Who/When/Where/Why military/military conflict/combatants./military/military combatant group/combatants Canada? \\\bottomrule
        
    \end{tabular}}
    \caption{The prompt we sue to verbalize the query triple.}
    \label{prompt1}
\end{table*}

We use FB15k-237 \citep{fb15k-237}, CoDEx-M \cite{codex}, and NELL-995 \citep{nell} for evaluation. FB15k-237 is a subset extracted from the Freebase \citep{free}, which includes commonsense knowledge about movies, sports, locations, etc. CoDEx-M is extracted from Wikipedia, which contains tens of thousands of hard negative triples, making it a more challenging KGC benchmark. NELL-995 is taken from the Never Ending Language Learner (NELL) system and covers many domains. Detailed statistics of all these datasets are shown in Table~\ref{dataset}.


\section{Baseline Details}\label{baseline}
\begin{table}[t]
    \centering
    \resizebox{\linewidth}{!}{
    \begin{tabular}{c|ccccc}\toprule
         \textbf{Dataset} & $\boldsymbol{\vert \mathcal{E} \vert}$ & $\boldsymbol{\vert \mathcal{R} \vert}$ & $\boldsymbol{\vert \textbf{Train} \vert}$ & $\vert \textbf{Valid} \vert$ & $\boldsymbol{\vert \textbf{Test} \vert}$  \\\midrule
         FB15k-237 & 14,541 & 237 & 272,115 & 17,535 & 20,466 \\
         CoDEx-M & 17,050 & 51 & 185,584 & 10,310 & 10,311 \\
         NELL-995 & 74,536 & 200 & 149,678 & 543 & 2,818
         \\\bottomrule
    \end{tabular}}
    \caption{Statistics of the Datasets.}
    \label{dataset}
\end{table}
\subsection{Baselines}
We compare our FtG against three types of baselines: (1) \textit{structure-based methods}, including TransE \citep{transe}, DistMult \citep{distmult}, ComplEx \citep{complex}, ConvE \citep{conve}, RotatE \citep{rotate}, and KG-Mixup \citep{KG-mixup}. (2) \textit{PLM-based methods}, including GenKGC \citep{genkgc}, KG-S2S \citep{kgs2s}, and CSProm-KG \citep{csform}. (3) \textit{LLM-based methods}, including ChatGPT \citep{chatgpt1shot}, PaLM2-540B \citep{Palm2}, and KG-LLaMA-7B \citep{kg-llama}. Both ChatGPT and PaLM2-540B employ LLMs as the backbone and focus on prompt engineer, enabling LLMs to understand the KGC task. KG-LLaMA-7B is an instruction fine-tuned LLaMA2-7B based on KG datasets.

\subsection{Implementation of Baselines}
Since some baselines miss results on some metrics, we implement these baselines based on their released code. For structure-based baselines, we use the toolkit provided in RotatE\footnote{\url{https://github.com/DeepGraphLearning/KnowledgeGraphEmbedding}}, which gives state-of-the-art performance of existing KGC models in a unified framework. We adopt the optimal hyperparameter configurations reported in their papers.

For KG-S2S, on the CoDEx-M dataset, we utilize the official code provided \footnote{\url{https://github.com/chenchens190009/KG-S2S}} and set the length of the entity description text to 30, epochs to 50, batch size to 32, and beam width to 40. For experiments on the NELL-995 dataset, the epoch is increased to 100, batch size is determined to be 64, and the learning rate is fixed at 0.001. It is worth noting that the NELL-995 dataset inherently lacks entity description texts, the length of entity description text is set to 0. For CSProm-KG \footnote{\url{https://github.com/chenchens190009/CSProm-KG}}, following the original paper, we choose ConvE as graph model owing to its superior performance metrics, alongside the BERT-base model serving as the foundational PLM. Specifically, on the CoDEx-M dataset, we set the batch size to 128, epoch to 60, length of entity description text to 30, learning rate to 0.0005, prompt length to 30, label smoothing to 0.1, and embedding dimension to 156. On the NELL-995 dataset, we adopt same configurations expect the length of entity description is set 0. For the KG-LLaMA-7B, we adopt same LoRA configurations in Table~\ref{lora} as our FtG for fair comparison. 

\section{Implementation Details}\label{implementation}
In our implementation, the quantity $k$ of candidates retained is selected from $\{10,20,30,40\}$. During training, we keep the RotatE frozen and employ LoRA to fine-tune the model. The detailed hyperparameters we use during training and inference are shown in Table~\ref{lora}. We employ identical hyperparameters in different datasets. DeepSpeed ZeRO stage3\footnote{\url{https://github.com/microsoft/Megatron-DeepSpeed}} is enabled for optimization. All models are trained using 2 Nvidia A800 GPUs, each with 80GB of memory. For all datasets, we report Mean Reciprocal Rank (MRR) and Hits@N (N=1,3,10) metric following the previous works.

\begin{table}[ht]
    \centering
    \resizebox{\linewidth}{!}{
    \begin{tabular}{c|c}
    \toprule
       \textbf{Name}  &  \textbf{Value} \\\midrule
        lora $r$ & 16 \\
        lora alpha & 32 \\
        lora dropout & 0.05 \\
        lora target modules & (q, k, v, o, down, up, gate) proj \\
        cutoff len & 1024 \\
        epochs & 2 \\
        per device batch size & 64 \\
        gradient accumulation steps & 1\\
        learning rate & $3e-4$ \\
        weight decay & $1e-5$ \\
        warm ratio & 0.01 \\
        lr scheduler type & cosine \\
        num return sequences & 10 \\
        projection layers & 1 \\\bottomrule
    \end{tabular}
    }
    \caption{Detailed hyperparameters used in our paper.}
    \label{lora}
\end{table}


\begin{table}[hb]
    \centering
    \resizebox{\linewidth}{!}{\begin{tabular}{c|c}
    \toprule
       \textbf{Hard negative candidates}  & \textbf{Explanation} \\\midrule
       (Senegal, part of, Middle East) & Senegal is part of West Africa. \\
       (Lesotho, official language, American English) & \makecell[c]{English, not American English, is an official \\ language of Lesotho.} \\
       (Vatican City, member of, UNESCO) & \makecell[c]{Vatican City is a UNESCO World Heritage \\ Site but not a member state.} \\
    \bottomrule
    \end{tabular}}
    \caption{Selected examples of hard samples in Codex.}
    \label{hard}
\end{table}


\section{Instruction Template}\label{ins}
\subsection{Prompt for ChatGPT and PaLM2}
\begin{figure}[ht]
    \centering
    \includegraphics[width=\linewidth]{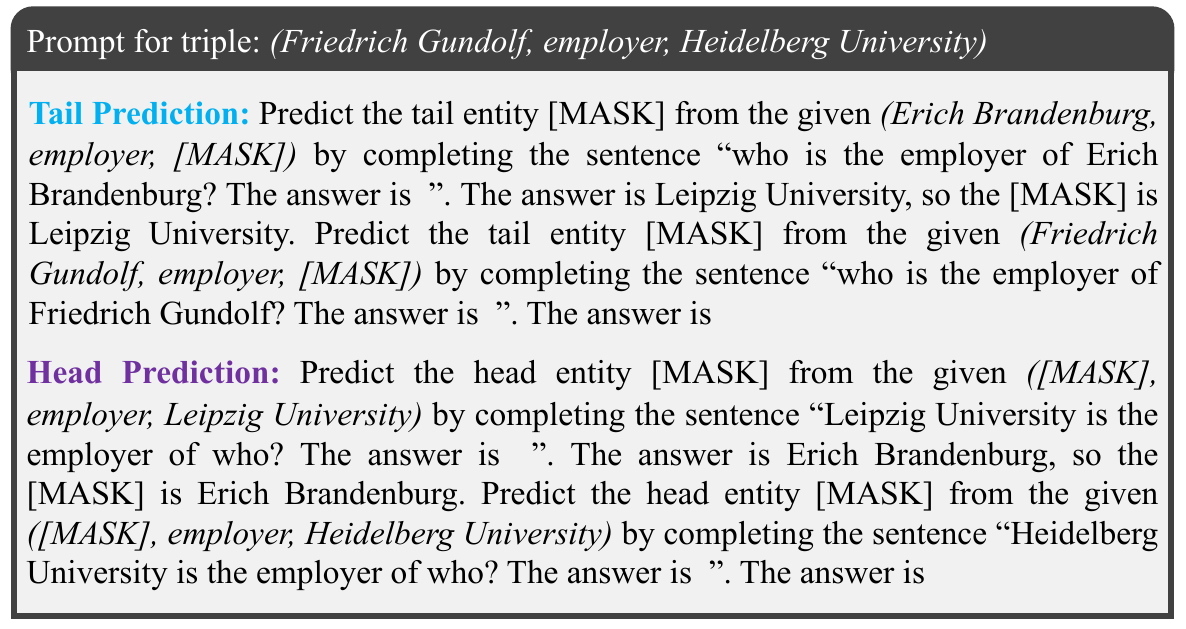}
    \caption{The prompt that directly leverage LLMs to perform KGC. Tail Prompt and Head Prompt mean the
input to predict the tail and head entity respectively.}
    \label{prompt_llm}
\end{figure}

\citet{chatgpt1shot} construct few-shot demonstrations to assess the performance of LLM in KGC. Figure~\ref{prompt_llm} shows a example of the input to LLMs, and \citep{distill} utilize the API parameter to obtain multiiple candidates, enabling the calculation of Hits@1, Hits@3, and Hits@10 metrics.

\subsection{Prompt for FtG} \label{our_prompt}
\begin{figure}[ht]
    \centering
    \includegraphics[width=\linewidth]{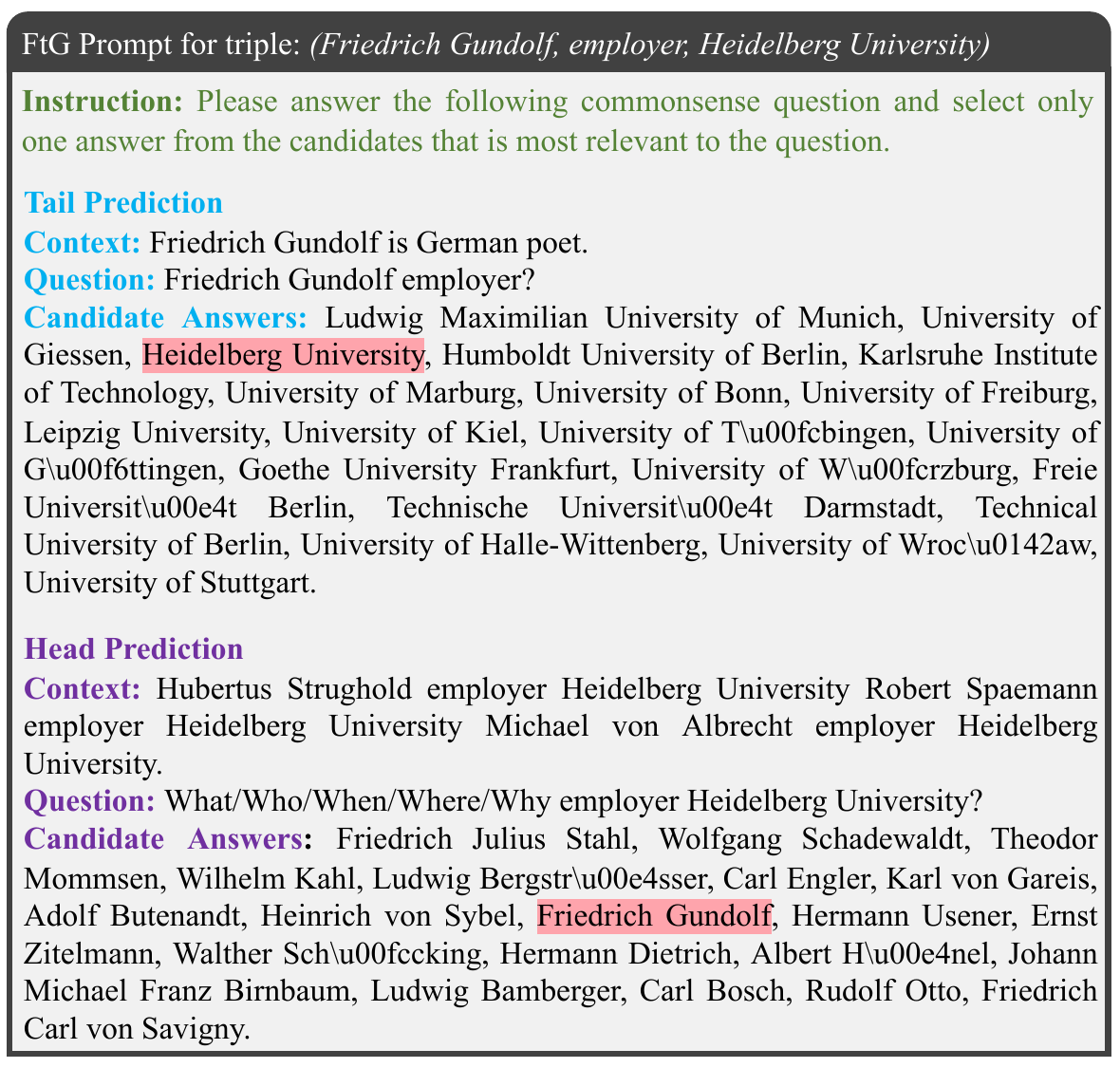}
    \caption{An example prompt that we adopt in FtG. \colorbox{pink}{\strut\strut} denotes target entity.}
    \label{prompt_ftg}
\end{figure}

In our framework, we use a simple template in Table~\ref{prompt1} to convert the query triple to text format like \citet{kg-llama}. Then we formulate the KGC task into a multiple-choice question fromat. A specific example is shown in Figure~\ref{prompt_ftg}.

\section{Case Study}
In Table~\ref{tab:my_label}, we demonstrate some cases to illustrate the differences in responses between existing LLMs-based methods and FtG. From the cases, our FtG can effectively leverage the capabilities of LLMs while avoiding the generation of uncontrollable text.
\begin{table}[ht]
    \centering
    \resizebox{\linewidth}{!}{\begin{tabular}{l|l|p{0.7\linewidth}} 
    \toprule
        \multicolumn{3}{l}{\textbf{Query head:} \textit{(?, employer, Heidelberg University)}} \\ 
    \midrule
        \textbf{Models} & \textbf{Labels} & \textbf{Output} \\
    \midrule
        LLaMA2-7B & \xmark & What/Who/When/Where/Why employer University of Cincinnati end{code} \\
    \midrule
        ChatGPT & \xmark & [MASK] is not able to provide a specific answer for the second sentence without more information.\\
    \midrule
        ChatGPT+FtG Prompt & \xmark & Hubertus Strughold, Robert Spaemann, and Michael von Albrecht are employed by Heidelberg University.\\ \midrule
        GPT4-o & \xmark & Max Weber. So, the [MASK] is Max Weber. \\\midrule
        GPT4-o+FtG Prompt & \cmark & Friedrich Gundolf \\\midrule
        FtG (ours) & \cmark & Friedrich Gundolf \\
    \bottomrule
    \end{tabular}}
    \resizebox{\linewidth}{!}{\begin{tabular}{l|l|p{0.7\linewidth}} 
    \toprule
        \multicolumn{3}{l}{\textbf{Query tail:} \textit{(Friedrich Gundolf, employer, ?)}} \\ 
    \midrule
        \textbf{Models} & \textbf{Labels} & \textbf{Output} \\
    \midrule
        LLaMA2-7B & \xmark &  Hermann Oncken (1879–1934) Theodor W. Adorno and Max Horkheimer, Dialectic of Enlightenment (1947) \\
    \midrule
        ChatGPT & \cmark & University of Heidelberg, so the [MASK] is University of Heidelberg.\\
    \midrule
        ChatGPT+FtG Prompt & \cmark & Heidelberg University\\ \midrule
        GPT4-o & \cmark & The answer is Heidelberg University. \\\midrule
        GPT4-o+FtG Prompt & \cmark & Heidelberg University \\\midrule
        FtG (ours) & \cmark & Heidelberg University \\
    \bottomrule
    \end{tabular}}
    \caption{Case comparisons between our FtG and existing LLMs. Corresponding prompt refer to Appendix~\ref{ins}.}
    \label{tab:my_label}
\end{table}

\section{Comparison with existing LLMs-based methods}\label{compare_llm}
In this subsection, we provide a detailed introduction to existing LLMs-based methods and further discuss the potential application of our FtG. The existing LLM-based methods mainly include:
\begin{itemize}
    \item \textbf{KoPA}~\citep{KoPA}: proposes an instruction-tuning method based on LLaMA2-7B for KG triple classification task. Although the authors claim their focus is on KGC, their work is strictly speaking a triple classification task. This means given a true triple, they randomly replace the head entity or tail entity to construct a negative sample, and then let the model perform binary classification, i.e., simply outputting True or False.

    \item \textbf{KG-LLM}~\citep{KG-LLM}: similarly focuses on the triple classification task. It constructs chain-of-thought prompts via random walks on the KG and then fine-tunes LLMs to perform binary classification. The results indicate that LLMs have a potential to understand structural information on the KG.

    \item \textbf{ChatGPT for KGC}~\citep{chatgpt1shot}: is the first approach that truly utilizes LLMs for KGC task, which converts the KGC task into sentence masking prediction. KICGPT \cite{kicgpt} focuses on enhancing the KGC by asking LLMs to perform re-ranking on preliminary results of a traditional KGC method.
    

    \item \textbf{Contextualization Distillation}~\citep{distill}: is a data-augmentation method that utilizes LLMs to generate background text about the query entity to enhance existing KGC methods that utilize textual information. However, due to the hallucination issues inherent to LLMs and the high requirements of downstream KGC methods, the improvement is limited.

    \item \textbf{KG-LLaMA}~\citep{kg-llama}: converts KGC into a QA task, leveraging the instruction-following capability of LLMs to adapt them for KGC tasks. While this approach has achieved promising results, it still falls short compared to previous methods based on structural information.
\end{itemize}
In contrast, our FtG can effectively leverage the capabilities of LLMs while avoiding the generation of uncontrollable text. Additionally, we propose an efficient approach to enable LLMs to utilize the structural information of KGs, which has not been achieved by previous methods.

\noindent\textbf{Potential Downstream Applications of FtG.} 
\begin{itemize}
    \item The construction of KG is very expensive, especially in specialized domains such as financial and law. Our method can fully leverage LLMs' inherent knowledge and reasoning capabilities, making it suitable for automatically completing existing KGs. Additionally, the AI community has witnessed the emergence of numerous powerful LLMs, which have made huge advancements and led to the pursuit of possible AGI. Our FtG provides a possible way to integrate KGs with LLMs, aligning with the current trends in AI domains.
    \item In the recommendation domain, systems need to suggest items from a vast pool. Our approach can be effectively applied here: filter-then-generate paradigm can initially filter the large pool of items, honing in on a more relevant subset based on user profiles and preferences. And the ego-graph serialization prompt can capture and model detailed user interaction history. Finally, the encoding the ego-graph into a soft prompt token and map it into LLMs' space with an adapter can provides a meaningful way to apply LLMs for final recommendation.
\end{itemize}

\end{document}